\pdfoutput=1
\documentclass[11pt]{article}

\usepackage[]{acl}

\usepackage{times}
\usepackage{latexsym}
\usepackage{enumitem}
\usepackage{float}
\usepackage{booktabs}
\usepackage{multirow, makecell}

\usepackage[T1]{fontenc}
\usepackage[utf8]{inputenc}

\usepackage{microtype}
\usepackage{caption}
\usepackage{subcaption}
\usepackage{makecell}
\usepackage{amsmath}

\newlength{\mysize}
\newcommand{\mycfs}[1]{\setlength{\mysize}{#1pt}%
  \fontsize{\mysize}{1.2\mysize}\selectfont}

\newcommand{\acpt}[1]{American Crossword Puzzle Tournament}

\newif\ifcomments
\ifcomments
    \providecommand{\eric}[1]{{\protect\color{magenta}{[EW: #1]}}}
    \providecommand{\albert}[1]{{\protect\color{magenta}{[AX: #1]}}}
    \providecommand{\eshaan}[1]{{\protect\color{magenta}{[EP: #1]}}}
    \providecommand{\nicholas}[1]{{\protect\color{magenta}{[NT: #1]}}}
    \providecommand{\kevin}[1]{{\protect\color{magenta}{[KY: #1]}}}
    \providecommand{\matt}[1]{{\protect\color{purple}{[MG: #1]}}}
    \providecommand{\dan}[1]{{\protect\color{purple}{[DK: #1]}}}
\else
    \providecommand{\eric}[1]{}
    \providecommand{\albert}[1]{}
    \providecommand{\eshaan}[1]{}
    \providecommand{\nicholas}[1]{}
    \providecommand{\kevin}[1]{}
    \providecommand{\matt}[1]{}
    \providecommand{\dan}[1]{}
\fi

\usepackage{graphicx}
\usepackage{booktabs}
\usepackage{tikz}
\usepackage{pifont}
\def\checkmark{\ding{52}} 
\def\xmark{\ding{55}}
\definecolor{gitred}{HTML}{FDB8C0}
\definecolor{gitgreen}{HTML}{ACF294}
\definecolor{darkred}{HTML}{8B0000}
\definecolor{eggshell}{HTML}{F5FAF6}

\newcommand{\xword}[1]{{\textsc{#1}}}

\newcommand{\edit}[1]{{{#1}}}

\usepackage{blindtext}
\renewenvironment{quote}{%
  \list{}{%
    \leftmargin0.3cm   % this is the adjusting screw
    \rightmargin\leftmargin
  }
  \item\relax
}
{\endlist}

\title{Automated Crossword Solving \vspace{0.3cm}}
\author{
Eric Wallace$^{\star}$ \\ UC Berkeley \And
Nicholas Tomlin$^{\star}$ \\ UC Berkeley \And
Albert Xu$^{\star}$ \\ UC Berkeley \And
Kevin Yang$^{\star}$ \\ UC Berkeley \AND
Eshaan Pathak$^{\star}$ \\ UC Berkeley \And
Matthew L. Ginsberg \\ Matthew Ginsberg, LLC \And
Dan Klein \\ UC Berkeley \\[0.9ex] {\hspace{-10cm}\{\href{mailto:ericwallace@berkeley.edu}{\tt ericwallace}, \href{mailto:nicholas\_tomlin@berkeley.edu}{\tt nicholas\_tomlin}, \href{mailto:albertxu3@berkeley.edu}{\tt albertxu3}, \href{mailto:klein@berkeley.edu}{\tt klein}\}\href{mailto:ericwallace@berkeley.edu}{\tt @berkeley.edu}}
}

\begin{document}
\maketitle
\begin{abstract}
We present the Berkeley Crossword Solver, a state-of-the-art approach for automatically solving crossword puzzles. Our system works by generating answer candidates for each crossword clue using neural question answering models and then combines loopy belief propagation with local search to find full puzzle solutions. Compared to existing approaches, our system improves exact puzzle accuracy from 71\% to 82\% on crosswords from \textit{The New York Times} and obtains 99.9\% letter accuracy on themeless puzzles.
% Our system also won first place at the top human crossword tournament, which marks the first time that a computer program has surpassed human performance at this event.
Additionally, in 2021, a hybrid of our system and the existing Dr.Fill system outperformed all human competitors for the first time at the American Crossword Puzzle Tournament.
To facilitate research on question answering and crossword solving, we analyze our system's remaining errors and release a dataset of over six million question-answer pairs.
\end{abstract}

\section{Introduction}

\begin{table*}
\centering
\small
\begin{tabular}{lllc}
\toprule
\textbf{Category} & \textbf{Clue} & \textbf{Answer} & \textbf{QA Recall} \\
\midrule
\multirow{2}{*}{Knowledge (37\%)} & Birds on Minnesota state quarters & \textsc{loons} & \checkmark \\[0.2ex]
 & Architect Frank & \textsc{gehry} & \checkmark \\
  \midrule 
  \multirow{2}{*}{Definition (33\%)} & First in a series & \textsc{pilot} & \checkmark \\[0.2ex]
  & Tusked savanna dweller & \textsc{warthog} & \checkmark \\
\midrule 
\multirow{2}{*}{Commonsense (14\%)} & Like games decided by buzzer beaters & \textsc{close} & \checkmark \\[0.2ex]
& Opposite of \textit{luego} & \textsc{ahora} & \checkmark \\
\midrule
\multirow{2}{*}{Wordplay (8\%)}
& Frequent book setting & \textsc{shelf} & \checkmark \\[0.2ex]
 & One followed by nothing? & \textsc{ten} & \xmark \\
\midrule
\multirow{2}{*}{Phrase (8\%)} & ``Is it still a date?'' & \textsc{areweon} & \checkmark \\[0.2ex]
 & ``Post \_\_\_ analysis'' & \textsc{hoc} & \checkmark \\
\midrule
\multirow{2}{*}{Cross-Reference (2\%)} & See Capital of 52-Down & \textsc{ghana} & \xmark \\[0.2ex]
& Oft-wished-upon sighting & \textsc{shootingmeteor} & \xmark \\
\bottomrule
\end{tabular}
\vspace{-0.15cm}
\caption{Types of reasoning used in \textit{The New York Times Crossword}. We compute each type's frequency by manually analyzing 200 clues. See Appendix~\ref{appendix:qualitative} for category definitions. We also indicate if our QA model correctly predicts each answer based on top-1000 recall.
Cross-reference clues mention other clues or themes, e.g., \textsc{shootingmeteor} replaces the clued phrase \textsc{shootingstar} based on the context from the puzzle.
}
\label{tab:clue_categories}
\end{table*}

\begin{quote}
\mycfs{11.5}
\vspace{-0.05cm}
\emph{``The key to solving crosswords is mental flexibility. If one answer doesn't seem to be working out, try something else.''}

--- Will Shortz, NYT Crossword Editor
\end{quote}

\noindent Crossword puzzles are perhaps the world’s most popular language game, with millions of solvers in the United States alone \cite{ginsberg2011dr}. Crosswords test knowledge of word meanings, trivia, commonsense, and wordplay, while also requiring one to simultaneously reason about multiple intersecting answers. Consequently, crossword puzzles provide a testbed to study open problems in AI and NLP, ranging from question answering to search and constraint satisfaction. In this paper, we describe an end-to-end system for solving crossword puzzles that tackles many of these challenges.

\begin{figure}
\centering
\includegraphics[width=0.9\linewidth]{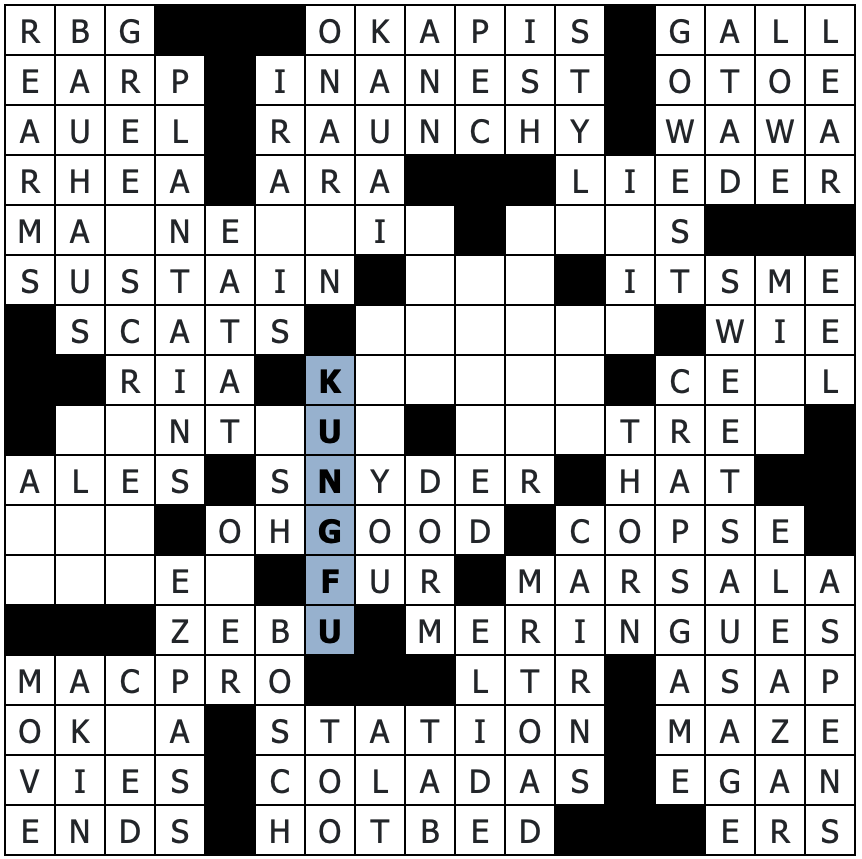}
\vspace{-0.1cm}
\caption{A partially-solved example crossword puzzle from the 2021 \acpt{}, where our system scored higher than all 1033 human solvers. The highlighted fill \textsc{kungfu} answers the wordplay clue: \textit{Something done for kicks?}}
\label{fig:crossword}
\end{figure}

\subsection{The Crossword Solving Problem}

Crossword puzzles are word games consisting of rectangular grids of squares that are to be filled in with letters based on given clues (e.g., Figure~\ref{fig:crossword}). Puzzles typically consist of 60–80 clues that vary in difficulty due to the presence of complex wordplay, intentionally ambiguous clues, or esoteric knowledge. Each grid cell belongs to two words, meaning that one must jointly reason about answers to multiple questions. Most players complete crosswords that are published daily in newspapers and magazines such as \textit{The New York Times} (NYT), while other more expert enthusiasts also compete in live events such as the \acpt{} (ACPT). These events are intensely competitive: one previous winner reportedly solved twenty puzzles per day as practice \cite{danfeyer}, and top competitors can perfectly solve expert-level puzzles with over 100 clues in just 3 minutes.

\noindent Automated crossword solvers have been built in the past and can outperform most hobbyist humans. Two of the best such systems are Proverb \citep{littman2002proverb} and Dr.Fill~\citep{ginsberg2011dr}. Despite their reasonable success, past systems struggle to solve the difficult linguistic phenomena present in crosswords, and they fail to outperform expert humans.
At the time of its publication, Proverb would have ranked 213th out of 252 in the ACPT.  Dr.Fill would have placed 43rd at publication and has since improved to place as high as 11th in the 2017 ACPT.

\subsection{A Testbed for Question Answering}
Answering crossword clues involves challenges not found in traditional question answering (QA) benchmarks. The clues are typically less literal; they span different reasoning types (c.f., Table~\ref{tab:clue_categories}); and they cover diverse linguistic phenomena such as polysemy, homophony, puns, and other types of wordplay. Many crossword clues are also intentionally underspecified, and to solve them, one must be able to ``know what they don't know'' and defer answering those clues until crossing letters are known.
Crosswords are also useful from a practical perspective as the data is abundant, well-validated, diverse, and constantly evolving. In particular, there are millions of question-answer pairs online, and unlike crowdsourced datasets that are often rife with artifacts~\cite{gururangan2018annotation,min2019compositional}, crossword clues are written and validated by experts. Finally, crossword data is diverse as it spans many years of pop culture, is written by thousands of different constructors, and contains various publisher-specific idiosyncrasies.

\subsection{A Testbed For Constraint Satisfaction}

Solving crosswords goes beyond just generating answers to each clue. Without guidance from a constraint solver, QA models cannot reconcile crossing letter and length constraints. Satisfying these constraints is challenging because the search space is enormous and many valid solutions exist, only one of which is correct. Moreover, due to miscalibration in the QA model predictions, exact inference may also lead to solutions that are high-likelihood but completely incorrect, similar to other types of structured decoding problems in NLP~\cite{nmtsearch,kumar2019calibration}. Finally, the challenges in search are amplified by the unique long tail of crossword answers, e.g., ``\emph{daaa bears}'' or ``\textit{eeny meeny miny moe},'' which makes it highly insufficient to restrict the search space to solutions that contain only common English words.

\begin{figure*}[h]
\centering
\includegraphics[trim={1.1cm 1cm 1.1cm 1.5cm},clip,width=\textwidth]{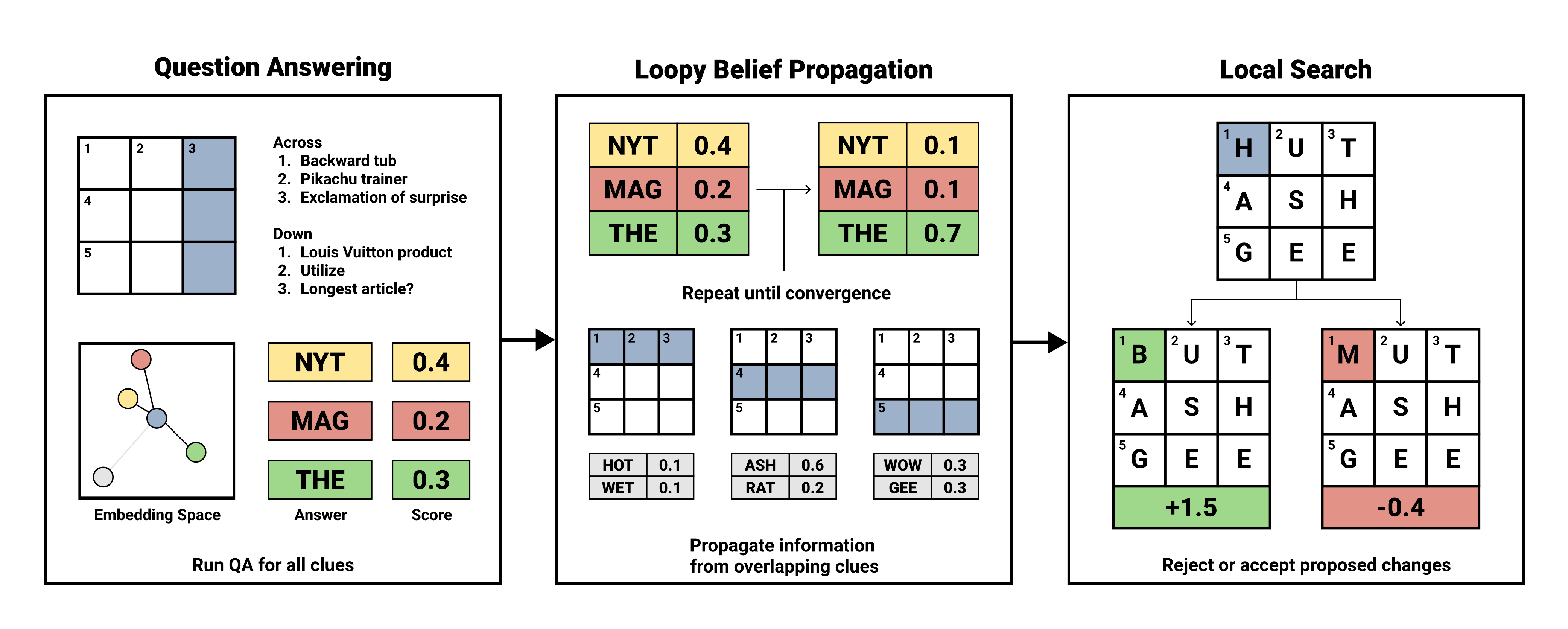}
\vspace{-0.65cm}
\caption{\textit{An overview of the Berkeley Crossword Solver.} We use a neural question answering model to generate answer probabilities for each question, and then refine the probabilities with loopy belief propagation. Finally, we fill the grid with greedy search and iteratively improve uncertain areas of the puzzle using local search.}
\label{fig:overview}
\end{figure*}

\subsection{The Berkeley Crossword Solver}

We present the Berkeley Crossword Solver (BCS), which is summarized in Figure~\ref{fig:overview}. The BCS is based on the principle that some clues are difficult to answer without any letter constraints, but other (easier) clues are more standalone. This naturally motivates a multi-stage solving approach, where we first generate answers for each question independently, fill in the puzzle using those answers, and then rescore uncertain answers while conditioning on the predicted letter constraints.
We refer to these stages as first-pass QA, constraint resolution, and local search, and we describe each component in Sections~\ref{sec:qa}--\ref{sec:reranking} after describing our dataset in Section~\ref{sec:dataset}. 
In Section~\ref{sec:results}, we show that the BCS substantially improves over the previous state-of-the-art Dr.Fill system, perfectly solving 82\% of crosswords from \textit{The New York Times}, compared to 71\% for Dr.Fill. Nevertheless, room for additional improvement remains, especially on the QA front. To facilitate further exploration, we publicly release our code, models, and dataset: \url{https://github.com/albertkx/berkeley-crossword-solver}.
\section{Crossword Dataset}\label{sec:dataset}

This section describes the dataset that we built for training and evaluating crossword solving systems. Recall that a crossword puzzle contains both question-answer pairs and an arrangement of those pairs into a grid (e.g., Figure~\ref{fig:crossword}). Unfortunately, complete crossword puzzles are protected under copyright agreements; however, their individual question-answer pairs are free-to-use. Our dataset efforts thus focused on collecting numerous question-answer pairs (Section~\ref{subsec:qa_pairs}) and we collected a smaller set of complete puzzle grids to use for final evaluation (Section~\ref{subsec:complete_puzzles}).

\subsection{Collecting Question-Answer Pairs}\label{subsec:qa_pairs} We collected a dataset of over six million question-answer pairs from top online publishers such as \textit{The New York Times}, \textit{The LA Times}, and \textit{USA Today}. We show qualitative examples in Table~\ref{tab:clue_categories}, summary statistics in Table~\ref{tab:statistics}, and additional breakdowns in Appendix~\ref{appendix:dataset}. Compared to existing QA datasets, our crossword dataset represents a unique and challenging testbed as it is large and carefully labeled, is varied in authorship, spans over 70 years of pop culture, and contains examples that are difficult for even expert humans.
We built validation and test sets by splitting off every question-answer pair used in the 2020 and 2021 NYT puzzles. We use recent NYT puzzles for evaluation because the NYT is the most popular and well-validated crossword publisher, and because using newer puzzles helps to evaluate temporal distribution shift.\medskip

\noindent \textbf{Word Segmentation of Answers} Crossword answers are canonically filled in using all capital letters and without spaces or punctuation, e.g., ``\emph{whale that stinks}'' becomes \textsc{whalethatstinks}. These unsegmented answers may confuse neural QA models that are pretrained on natural English text that is tokenized into wordpieces. To remedy this, we trained a word segmentation model that maps the clues to their natural language form.\footnote{More simplistic algorithms that segment the answer into known English words are insufficient for many crossword answers, e.g., \textsc{daaabears} and \textsc{eenymeenyminymoe}.} We collected segmentation training data by retrieving common $n$-grams from Wikipedia and removing their spaces and punctuation. We then finetuned GPT-2 small~\cite{radford2019gpt2} to generate the segmented $n$-gram given its unsegmented version. We ran the segmenter on all answers in our data. In all our experiments, we train our QA models using segmented answers and we post-hoc remove spaces and punctuation from their predictions.

\subsection{Collecting Complete Crossword Puzzles}\label{subsec:complete_puzzles}
To evaluate our final crossword solver, we collected a validation and test set of complete 2020 and 2021 puzzle grids. We use puzzles from \textit{The New York Times}, \textit{The LA Times}, \textit{Newsday}, \textit{The New Yorker}, and \textit{The Atlantic}.
Using multiple publishers for evaluation provides a unique challenge as each publisher contains different idiosyncrasies, answer distributions, and crossword styles. We use 2020 NYT as our validation set and hold out all other puzzles for testing. There are 408 total test puzzles.

% \eshaan{self-reminder 1: rename all of the files in our dataset from "valid" to "test".} 
% \eshaan{self-reminder 2: update our dataset release for segmented answers with wordsegment library?}

\begin{table}[t]
\begin{center}
\setlength{\tabcolsep}{4pt}
\begin{tabular}{lccc}
\toprule
 & \textbf{Train} &  \textbf{Validation} & \textbf{Test} \\
\midrule
QA Pairs & 6.4M & 30.4K & 21.3K \\
Answer Set & 437.8K & 17.2K & 13.4K \\
% \# Puzzles & 122.2K & 366 & 255 \\
% \# Unique Sources & 27 & 1 & 1 \\
Timeframe & 1951-2019 & 2020 & 2021 \\
% \# Puzzles & 122,176 & 366 & 255 \\
% \# Unique Sources & 27 & 1 & 1 \\
% \# QA Pairs & 6.44M & 30,421 & 21,301 \\
% \# Unique Answers & 437,770 & 17,246 & 13,350 \\
% \% Answers in Train & 100 & 97.0 & 96.6 \\
\bottomrule
\end{tabular}
\end{center}
\vspace{-0.3cm}
\caption{\textit{Summary statistics of our QA dataset.} We collect question-answer pairs from 26 sources (\textit{The LA Times}, \textit{The New York Times}, etc.) for training, and we hold out the latest data from NYT for validation and testing. Our dataset is large and contains a wide range of authors, answers, puzzle sources, and years.}
\label{tab:statistics}
\end{table}
\section{Bi-Encoder QA Model}\label{sec:qa}

The initial step of the BCS is question answering: we generate a list of possible answer candidates and their associated probabilities for each clue.
A key requirement for this QA model is that it does not output unreasonable or overly confident answers for hard clues. Instead, this model is designed to be used as a ``first-pass'' that generates reasonable candidates for every clue, in hope that harder clues can be reconciled later when predicted letter constraints are available. 
We achieve this by restricting our first-pass QA model to only output answers that are present in the training set. As discussed in Section~\ref{sec:reranking}, we later generate answers outside of this closed-book set with our second-pass QA model. 

\paragraph{Model Architecture} We build our QA model based on a bi-encoder architecture~\cite{siamese,karpukhin2020dense} due to its ability to score numerous answers efficiently and learn using few examples per answer.
We have two neural network encoders: $\mathrm{E}_{\mathrm{C}}(\cdot)$, the clue encoder, and $\mathrm{E}_\mathrm{A}(\cdot)$, the answer encoder.
Both encoders are initialized with BERT-base-uncased~\cite{devlin2018BERT} and output the encoder's [CLS] representation as the final encoding. These two encoders are trained to map the questions and answers into the same feature space. Given a clue $c$, the model scores all possible answers $a_i$ using a dot product similarity function between feature vectors: $\mathrm{sim}(c, a_i) = \mathrm{E}_\mathrm{C}(c)^{\text{T}} \mathrm{E}_\mathrm{A}(a_i)$. Our answer set consists of the 437.8K answers in the training data.\footnote{Our bi-encoder model is a ``closed-book'' QA model because it does not have ``open-book'' access to external knowledge sources such as Wikipedia~\cite{roberts2020much}. We found in preliminary experiments that open-book models struggle as most crossword answers are not present or are difficult to retrieve from knowledge sources such as Wikipedia.}\medskip

\noindent \textbf{Training} We train the encoders in the same fashion as DPR~\cite{karpukhin2020dense}: batches consist of clues, answers, and ``distractor'' answers. The two encoders are trained jointly to assign a high similarity to the correct question-answer pairs and low similarity to all other pairs formed between the clue and distractor answers. We use one distractor answer per clue that we collect by searching each clue in the training set using TFIDF and returning the top incorrect answer. We tune hyperparameters of our bi-encoder model based on its top-$k$ accuracy on the NYT validation set.\medskip

\noindent \textbf{Inference} At test time, for each clue $c$, we compute the embedding $v_c = E_C(c)$ and retrieve the answers whose embeddings have the highest dot product similarity with $v_c$. We obtain probabilities for each answer by softmaxing the dot product scores. To speed up inference, we precompute the answer embeddings and use FAISS~\cite{johnson2019billion} for similarity scoring.

\subsection{Top-k Recall of Our QA Model}
\begin{figure}[t]
\centering
\includegraphics[trim={0cm 0cm 0cm 0cm},clip,width=1.0\columnwidth]{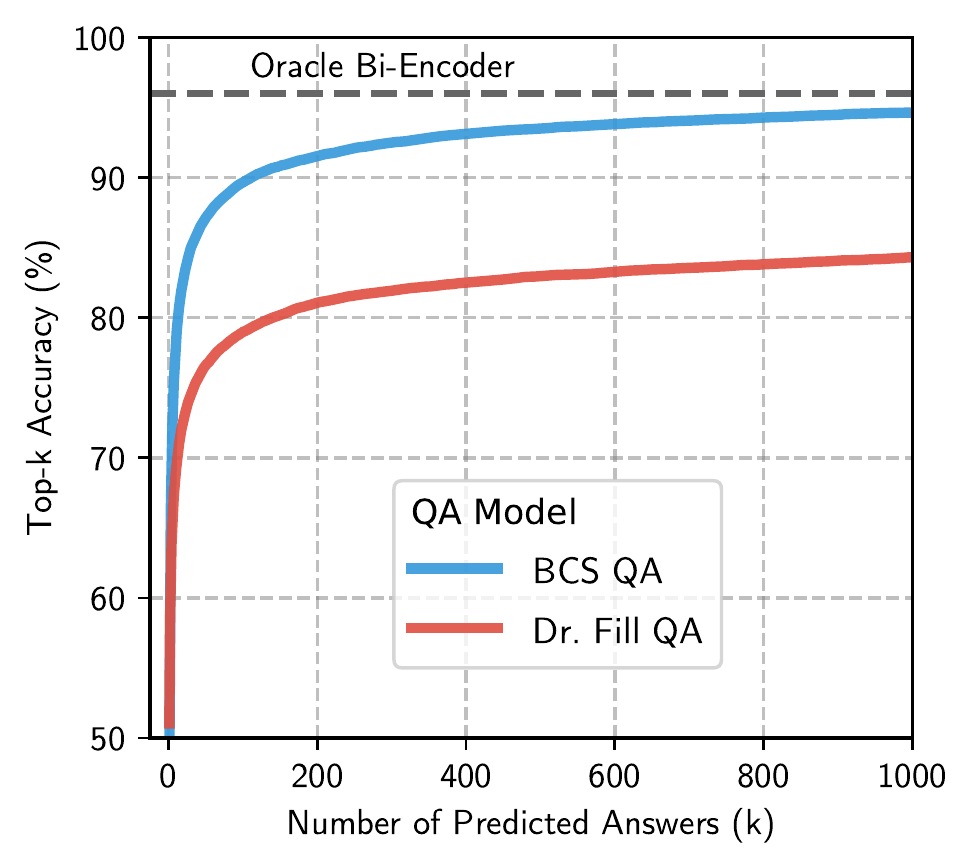}
\vspace{-0.73cm}
\eshaan{can we capitalize the first "e" in bi-encoder in the plot? for consistency}
\caption{\textit{Top-$k$ accuracy on the 2021 NYT test set}. Dr.Fill QA is an existing crossword QA system that ensembles TFIDF-like scoring with numerous additional scoring modules. Our neural bi-encoder model improves top-1000 accuracy from 84.4\% to 94.6\%.}
\label{fig:topk}
\end{figure}

To evaluate our bi-encoder, we compute its top-$k$ recall on the question-answer pairs from the NYT test set. We are most interested in top-1000 recall, as we found it to be highly-correlated with downstream solving performance (discussed in Section~\ref{sec:erroranalysis}).
As a baseline, we compare against the QA portion of the previous state-of-the-art Dr.Fill crossword solver~\cite{ginsberg2011dr}. This QA model works by ensembling TFIDF-like scoring and numerous additional modules (e.g., synonym matching, POS matching). Our bi-encoder model considerably outperforms Dr.Fill, improving top-1000 recall from 84.4\% to 94.6\% (Figure~\ref{fig:topk}). 
Also note that approximately 4\% of test answers are not seen during training, and thus the oracle recall for our first-pass QA model is $\approx$ 96\%.

\section{Resolving Letter Constraints Using BP}\label{section:contraints}

Given the list of answer candidates and their associated probabilities from the first-pass QA model, we next built a solver that produces a puzzle solution that satisfies the letter constraints. Formally, crossword solving is a weighted constraint satisfaction problem, where the probability over solutions is given by the product of the confidence scores produced by the QA model~\cite{ginsberg2011dr}. There are numerous algorithms for solving such problems, including branch-and-bound, integer linear programming, and more. 

We use belief propagation~\cite{PEARL1988143}, henceforth BP, for two reasons. First, BP directly searches for the solution with the highest \textit{expected overlap} with the ground-truth solution, rather than the solution with the highest likelihood under the QA model~\cite{littman2002proverb}. This is advantageous as it maximizes the total number of correct words and letters in the solution, and it also avoids strange solutions that may have spuriously high scores under the QA model. Second, BP also produces marginal distributions over words and characters, which is useful for generating an $n$-best list of solution candidates (used in Section~\ref{sec:reranking}).\medskip

\noindent \textbf{Loopy Belief Propagation} We use loopy BP, inspired by the Proverb crossword solver~\cite{littman2002proverb}. That is, we construct a bipartite graph with nodes for each of the crossword's clues and cells. For each clue node, we connect it via an edge to each of its associated cell nodes (e.g., a 5-letter clue will have degree 5 in the constructed graph). Each clue node maintains a belief state over answers for that clue, which is initialized using a mixture of the QA model's probabilities and a unigram letter LM.\footnote{The unigram letter LM accounts for the probability that an answer is not in our answer set. We build the LM by counting the frequency of each letter in our QA training set.} Each cell node maintains a belief state over letters for that cell. We then iteratively apply BP with each iteration doing message passing for all clue nodes in parallel and then for all cell nodes in parallel. The algorithm empirically converges after 5--10 iterations and completes in just 10 seconds on a single-threaded Python process.\medskip

\noindent \textbf{Greedy Inference} BP produces a marginal distribution over words for each clue. To generate an actual puzzle solution, we run greedy search where we first fill in the answer with the highest marginal likelihood, remove any crossing answers that do not share the same letter, and repeat.

\section{Iteratively Improving Puzzle Solutions}\label{sec:reranking}

\begin{figure*}[t]
\begin{subfigure}{0.235\textwidth}
\centering
\includegraphics[trim={0cm 0.15cm 0cm 0.4cm},clip,width=1.0\textwidth]{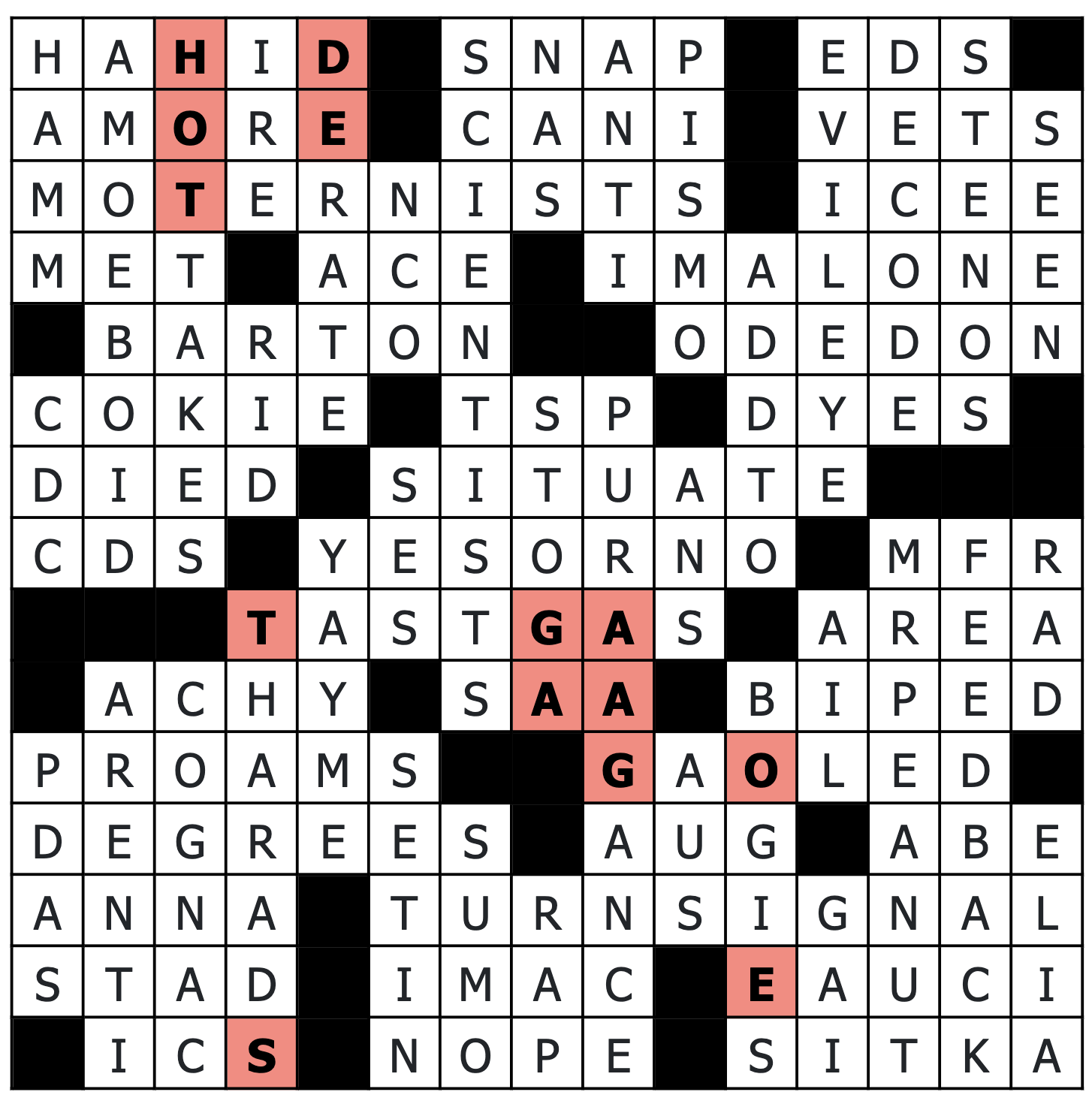}
\caption{Before Local Search}
\end{subfigure}%
\hfill
\begin{subfigure}{0.235\textwidth}
\centering
\includegraphics[trim={0cm 0.35cm 0cm 0.05cm},clip,width=1.0\textwidth]{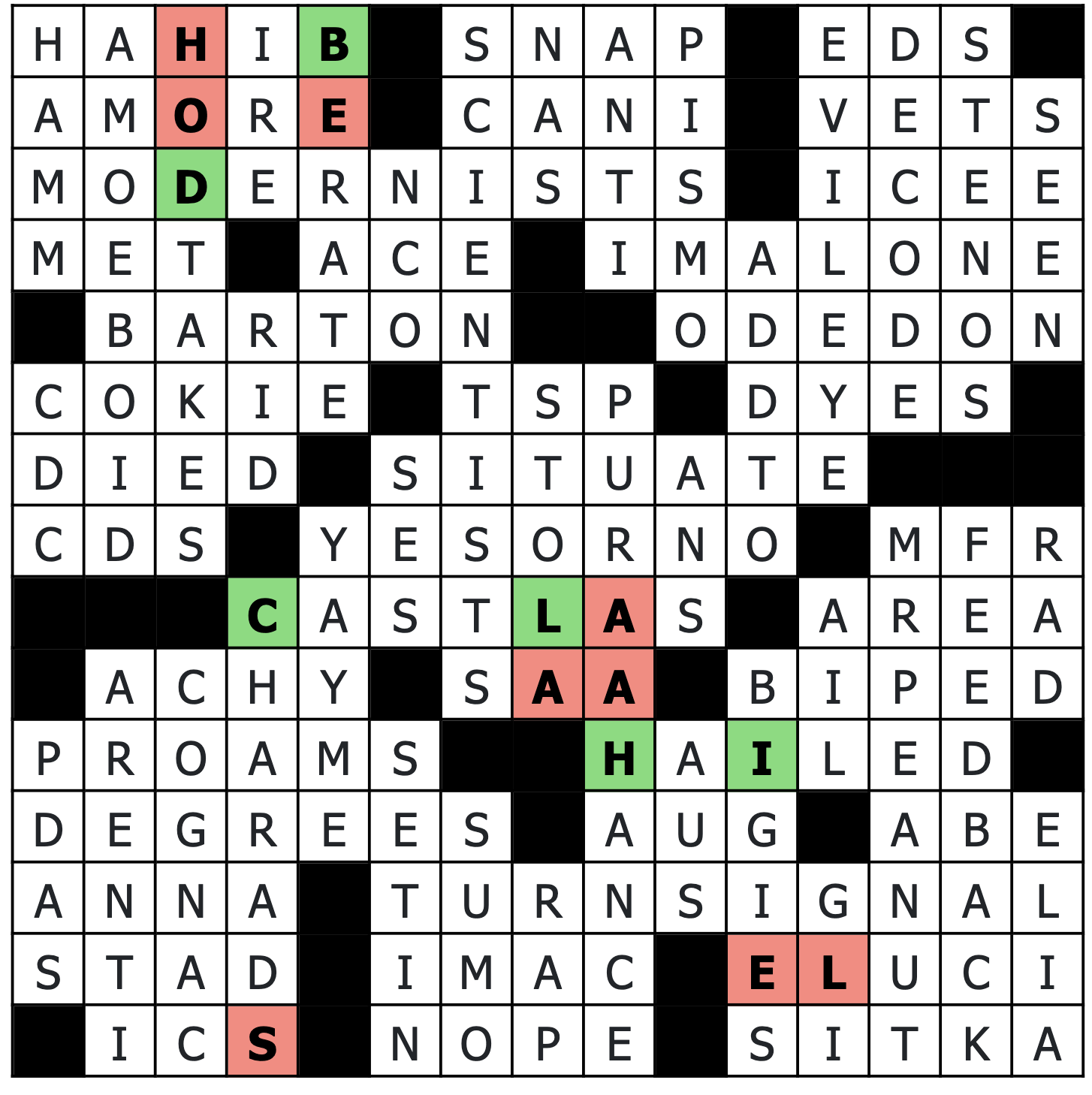}
\caption{Step \#1}
\end{subfigure}%
\hfill
\begin{subfigure}{0.235\textwidth}
\centering
\includegraphics[trim={0cm 0.1cm 0cm 0.25cm},clip,width=1.0\textwidth]{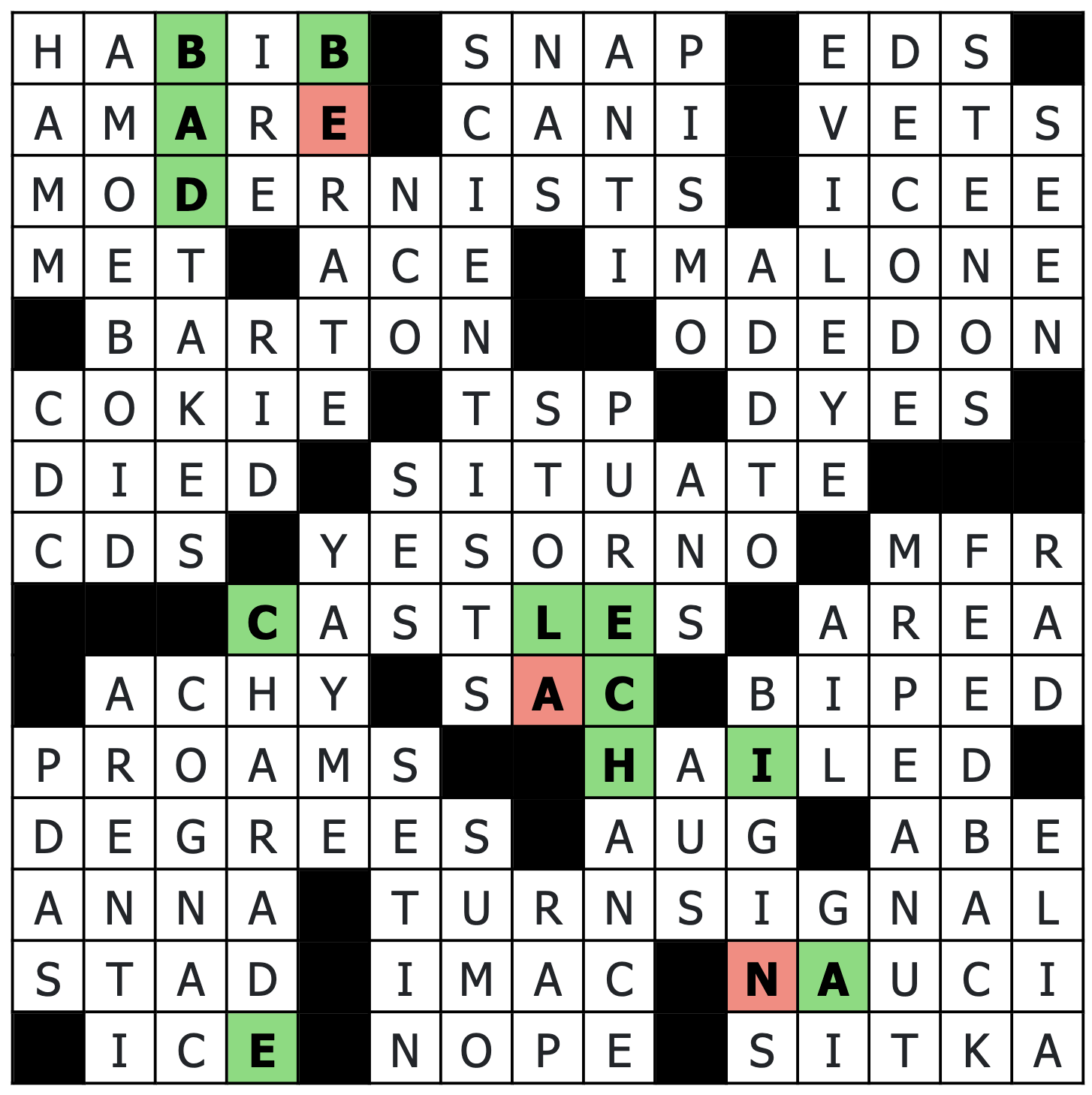}
\caption{Step \#2}
\end{subfigure}%
\hfill
\begin{subfigure}{0.235\textwidth}
\centering
\includegraphics[trim={0cm 0.3cm 0cm 0.1cm},clip,width=1.0\textwidth]{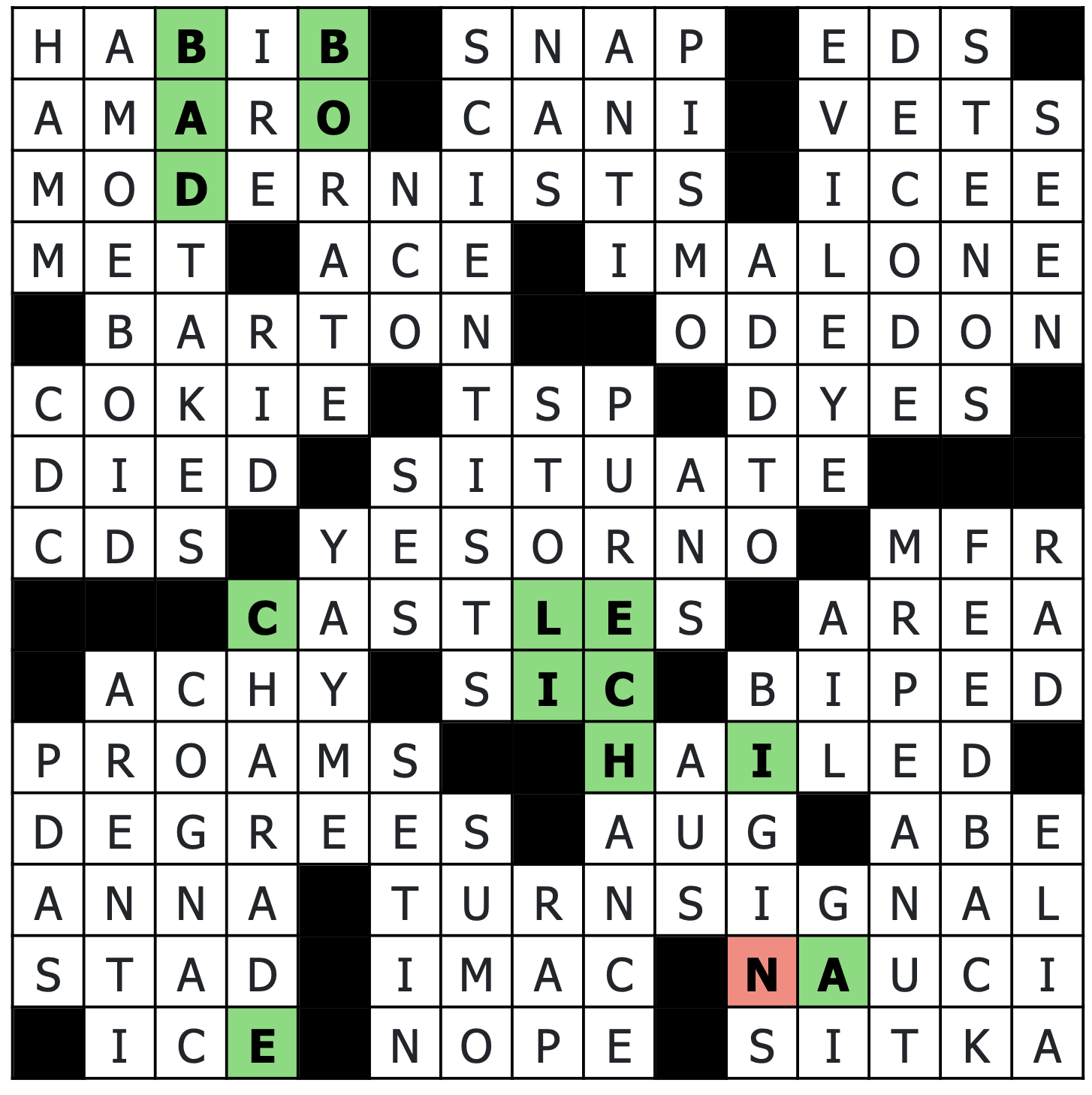}
\caption{Step \#3}
\end{subfigure}%
\vspace{-0.2cm}
\caption{We show the result of our solver on a NYT puzzle after running greedy search and three consecutive steps of local search. Local search considerably improves accuracy but fails to fix the answer regarding Dr. Fauci (an error due to temporal shift in our QA models). Red squares indicate errors from the output of greedy search, while green squares indicate corrections from the local search. See Figure~\ref{tab:ii_details} for the clues and associated answers in the puzzle.}
\label{fig:ii_example}
\end{figure*}

Many of the puzzle solutions generated by BP are close to correct but have small letter mistakes, e.g., \textsc{nauci} instead of \textsc{fauci} or \textsc{tazoambassadors} instead of \textsc{jazzambassadors}, as shown in Figure~\ref{fig:ii_example}.\footnote{These errors stem from multiple sources. First, 4\% of the answers in a test crossword are not present in our bi-encoder's answer set. Those answers will be not be filled in correctly unless the solver can identify the correct answer for \textit{all} of the crossing answers. Second, natural QA errors exist even on questions with non-novel answers. Finally, the BP algorithm may converge to a sub-optimal solution.}
We remedy this in the final stage of the BCS with local search (LS), where we take a ``second-pass'' through the puzzle and score alternate proposals that are a small edit distance away from the BP solution. In particular, we alternate between proposing new candidate solutions by flipping uncertain letters and scoring those proposals using a second-pass QA model.

\paragraph{Proposing Alternate Solutions} Similar to related problems in structured prediction~\cite{nmtsearch} or model-based optimization~\cite{fu2021offline}, the key challenge in searching for alternate puzzle solutions is to avoid false positives and adversarial inputs.
If we score \textit{every} proposal within a small edit distance to the original, we are bound to find nonsensical character flips that nevertheless lead to higher model scores.
We avoid this by only scoring proposals that are within a 2-letter edit distance and also have nontrivial likelihoods according to BP or a dictionary.
Specifically, we score all proposals whose 1--2 modified letters each have probability 0.01 or greater under the character marginal probabilities produced by BP.\footnote{The character-level marginal distribution for most characters assigns all probability mass to a single letter after a few iterations of BP (e.g., probability 0.9999). We empirically chose 0.01 as it achieved the highest validation accuracy.} We also score all proposals whose 1--2 modified letters cause the corresponding answer to segment into valid English words.\footnote{For instance, given a puzzle that contains a fill such as \textsc{munnyandclyde}, we consider alternate solutions that contain answers such as \textsc{bunnyandclyde} and \textsc{sunnyandclyde}, as they segment to ``\emph{bunny and clyde}'' and ``\emph{sunny and clyde}.''}

\paragraph{Scoring Solutions With Second-Pass QA} Given the alternate puzzle solutions, we could feed each of them into our bi-encoder model for scoring. However, we found that bi-encoders are not robust---they sometimes produce high-confidence predictions for the nonsensical answers present in some candidate solutions. We instead use generative QA models to score the proposed candidates as we found these models to be empirically more robust. We finetuned the character-level model ByT5-small~\cite{xue2021byt5} on our training set to generate the answer from a given clue. We then score each proposed candidate using the product of the model's likelihoods of the answers given the clues, $\prod_{j} P(a_j\mid c_j)$. 

After scoring all candidate proposals, we apply the best-scoring edit and repeat the proposal and scoring process until no better edits exist. Figure~\ref{fig:ii_example} shows an example of the candidates accepted by LS. Quantitatively, we found that LS applied 243 edits that improved accuracy and 31 edits that hurt accuracy across 234 NYT test puzzles.

\begin{table*}[h]
\centering
\begin{tabular}{lccccccc}
\toprule
 & & \multicolumn{2}{c}{Perfect Puzzle (\%)} & \multicolumn{2}{c}{Word Acc. (\%)}  & \multicolumn{2}{c}{Letter Acc. (\%)} \\
      Source & \# Puzzles & Dr. Fill & BCS & Dr. Fill & BCS & Dr. Fill & BCS \\
\midrule
{The Atlantic} & 46 & \edit{82.6} & \textbf{89.1} & 98.5 & \textbf{99.1} & 99.7 & \textbf{99.8} \\
{Newsday} & 52 & \edit{86.2} & \textbf{94.2} & 98.6 & \textbf{99.6} & 99.1 & \textbf{99.8} \\
{The New Yorker} & 22 & \textbf{\edit{86.4}} & 77.2 & \textbf{99.5} & 98.9 & \textbf{99.9} & 99.8 \\
%\midrule
{The LA Times} & 54 & \edit{81.5} & \textbf{92.6} & 99.4 & \textbf{99.7} & 99.9 & 99.9 \\
{The New York Times} & 234 & \edit{70.5} & \textbf{81.7} & \edit{97.9} & \textbf{98.9} & \edit{99.2} & \textbf{99.7} \\
\bottomrule
\end{tabular}
\vspace{-0.1cm}
\caption{\emph{Final results of the Berkeley Crossword Solver.} We compare the BCS to Dr. Fill, the previous state-of-the-art crossword solving system, on a range of puzzle sources. The BCS produces significantly more perfect puzzles and achieves better or comparable letter-level and word-level accuracies.}
\label{tab:results}
\end{table*}
\section{End-to-End System Results}\label{sec:results}

We evaluate our final system on our set of test puzzles and compare the results to the state-of-the-art Dr.Fill system~\cite{ginsberg2011dr}.
% \footnote{Note that while the original Dr.Fill paper was published in 2011, the system has been consistently updated and has substantially improved. Dr.Fill can outperform all but the best human solvers (see Table~\ref{tab:drfill_acpt} for statistics on its improvement).We run the latest system.}
We compute three accuracy metrics: perfect puzzle, word, and letter. Perfect puzzle accuracy requires answering every clue in the puzzle correctly and serves as our primary---and most challenging---metric.

Table~\ref{tab:results} shows our main results.
We outperform Dr.Fill on perfect puzzle accuracy across crosswords from every publication source. For example, we obtain a 11.2\% absolute improvement on perfect puzzle accuracy on crossword puzzles from \textit{The New York Times}, which is a statistically significant improvement ($p<0.01$) according to a paired $t$-test. We also observe comparable or better word and letter accuracies than Dr.Fill across all sources.
Our improvement on puzzles from \textit{The New Yorker} is relatively small; this discrepancy is possibly due to the small amount of data from \textit{The New Yorker} in our training set (see Figure~\ref{fig:publishers}). 
\paragraph{Themed vs. Themeless Puzzles} Although the BCS achieves equivalent or worse letter accuracy on \textit{Newsday} and \textit{LA Times} puzzles, it obtains substantially higher puzzle accuracy on these splits. We attribute this behavior to errors concentrated in unique themed puzzles, e.g., ones that place multiple letters into a single cell.
To test this, we break down NYT puzzles into those with and without special theme entries (see Appendix~\ref{appendix:analysis} for our definition of theme puzzles). On themeless NYT puzzles, we achieve 99.9\% letter accuracy and 89.5\% perfect puzzles, showing that themed puzzles are a major source of our errors. Note that the Dr.Fill system includes various methods to detect and resolve themes and is thus more competitive on such puzzles, although it still underperforms our system.

\paragraph{\acpt{}}
For our last evaluation, we submitted a system to participate \emph{live} in the \acpt{} (ACPT), the longest-running and most prestigious human crossword tournament. Our team obtained special permission from the organizers to participate in the 2021 version of the tournament, along with 1033 human competitors. For the live tournament, we used an earlier system, which does not use belief propagation or local search but instead uses Dr.Fill's constraint-resolution system along with the BCS QA modules described above. The submitted system outperformed all of the human participants --- we had a total score of 12,825 compared to the top human who had 12,810 (scoring details in Appendix~\ref{appendix:acpt}). Figure~\ref{fig:acpt_2021} shows our scores compared to the top and median human competitor on the 7 puzzles used in the competition. We also \emph{retrospectively} evaluated the final BCS system as detailed in this paper (i.e., using our solver based on belief propagation and local search), and achieved a higher total score of 13,065. This corresponds to getting 6 out of the 7 puzzles perfect and 1 letter wrong on 1 puzzle.

\paragraph{System Ablations}
We also investigated the importance of our QA model, BP inference, and local search with an ablation study. Table~\ref{tab:ablation} shows results for perfect puzzle accuracy on NYT 2021 puzzles under different settings.
The first ablation shows that our local search step is crucial for our solver to achieve high accuracy.
The second and third ablations show that the BCS's QA and solver are both superior to their counterparts from Dr.Fill---swapping out either component hurts accuracy. 

\eric{what does our system + Dr.Fill mean. We need to talk about how we ensemble his QA with ours.}

\begin{table}[t]
\centering
\begin{tabular}{lc}
\toprule
\textbf{System} & \textbf{Puzzle (\%)} \\
\midrule
BCS QA + BP + LS & 81.7\\
\midrule
BCS QA + BP & 44.3 \\
BCS QA + Dr.Fill Solver & 73.7 \\
Dr.Fill QA + Dr.Fill Solver & 70.5 \\
\bottomrule
\end{tabular}
\vspace{-0.1cm}
\caption{\emph{Ablations on NYT puzzles.} Our full system consists of a bi-encoder QA model, loopy belief propagation (BP), and local search (LS). We find that our QA and solver are both superior to that of Dr.Fill and that our local search step is key to achieving high accuracy.}
\label{tab:ablation}
\end{table}

\eshaan{we should state why we can't have our QA + LS to compare results to}

\begin{figure*}[t]
\centering
\includegraphics[trim={0cm 0cm 0cm 0cm},clip,width=\textwidth]{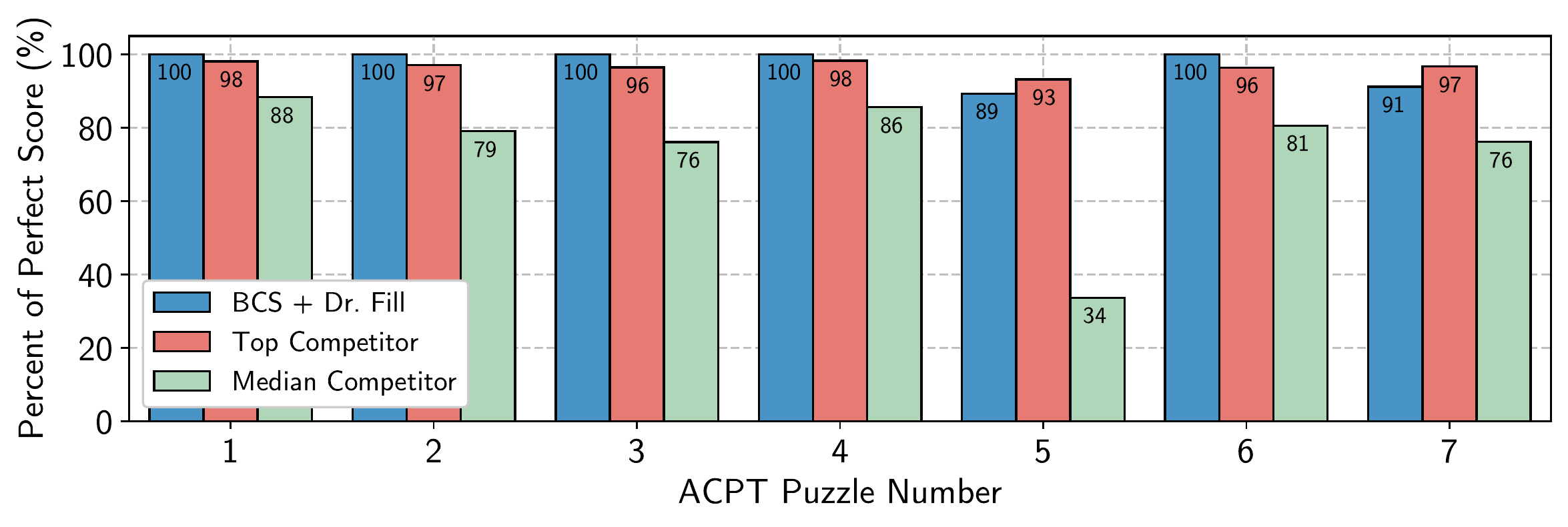}
\vspace{-0.75cm}
\caption{\textit{A breakdown of our 2021 ACPT performance.} The 2021 ACPT consisted of 7 puzzles, for which our combined system achieves a perfect score and surpasses the top human competitor on 5 out of the 7 puzzles. We include the median competitor's performance to illustrate the difficulty of the puzzles.}
\label{fig:acpt_2021}
\end{figure*}
\newpage\section{Error Analysis}\label{sec:erroranalysis}
% Our system outperforms the best human solvers; does this mean that crosswords are solved? 
Although our system obtains near-perfect accuracy on a wide variety of puzzles, we maintain that crosswords are not yet solved. In this section, we show that substantial headroom remains on QA accuracy and the handling of themed puzzles.
% The answer is, of course, no. In this section, we show that substantial headroom remains on QA accuracy and the handling of special themed puzzles.

\begin{figure}[t]
\centering
\includegraphics[trim={0.2cm 0cm 0.5cm 0cm},width=\linewidth]{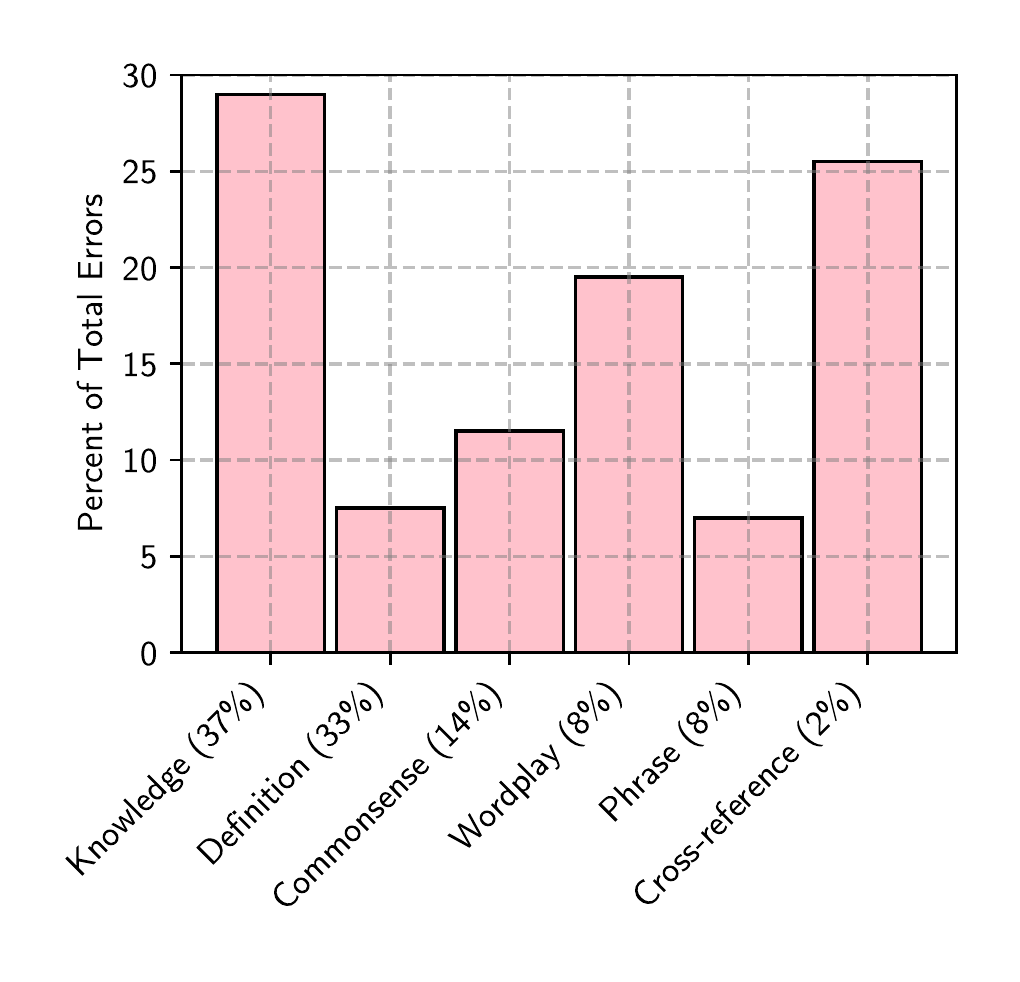}
\vspace{-1.1cm}
\caption{We manually categorize our QA failures using the categories from Table~\ref{tab:clue_categories}. The rate at which each category occurs in random examples is shown in parentheses. A disproportionate fraction of QA errors are due to cross-reference and wordplay clues.}
\label{fig:failure_categorization}
\end{figure}

\paragraph{QA Error Analysis} We first measured how well a QA model needs to perform on each clue in order for our solver to find the correct solution. We found that when our QA model ranks the true answer within the top 1,000 predictions, the answer is almost always filled in correctly (Figure~\ref{fig:unk_words_accuracy}). 
Despite top-1000 accuracy typically being sufficient, our QA model still makes numerous errors.
We manually analyzed these mistakes 
%in two ways. First, we break down accuracy as a function of answer length (Figure~\ref{fig:qa_acc_vs_fill_len}). Accuracy decreases monotonically with length, likely because longer answers are often more difficult multi-word expressions (e.g., \textsc{eenymeenyminymoe}). 
by sampling 200 errors from the NYT 2021 puzzles and placing them in the same categories used in Table~\ref{tab:clue_categories}. 
%Each clue is annotated by three of the authors and the final category is decided by majority vote. 
Figure~\ref{fig:failure_categorization} shows the results and indicates that knowledge, wordplay, and cross-reference clues make up the majority of errors.

\paragraph{End-to-end Analysis} We next analyzed the errors for our full system. There are 43 NYT 2021 puzzles that we did not solve perfectly. We manually separated these puzzles into four categories:
\begin{itemize}[leftmargin=10pt,itemsep=0mm, topsep=3pt]
    \item \textbf{Themes} (21 puzzles). Puzzles with unique themes, e.g., placing four characters in one cell.
    \item \textbf{Local Search Proposals} (9 puzzles). Puzzles where we did not propose a puzzle edit in local search that would have improved accuracy. 
    \item \textbf{Local Search Scoring} (9 puzzles). Puzzles where the ByT5 scorer either rejected a correct proposal or accepted an incorrect proposal. 
    \item \textbf{Connected Errors} (4 puzzles). Puzzles with errors that cannot be fixed by local search, i.e., there are several connected errors. 
\end{itemize}
\noindent Overall, the largest source of remaining puzzle failures is special themed puzzles, which is unsurprising as the BCS system does not explicitly handle themes. The remaining errors are mostly split between proposal and scoring errors. Finally, connected errors typically arise when BP fills in an answer that is in our bi-encoder's answer set but is incorrect, i.e., the first-pass model was overconfident.

\section{Related Work}\label{sec:related}

\noindent \textbf{Past Crossword Solvers}
Prior to our work, the three most successful automated crossword solvers were Proverb, WebCrow \citep{ernandes2005webcrow}, and Dr.Fill. Dr.Fill uses a relatively straightforward TFIDF-like search for question answering, but Proverb and WebCrow combine a number of bespoke modules for QA; WebCrow also relies on a search engine to integrate external knowledge. On the solving side, Proverb and WebCrow both use loopy belief propagation, combined with A* search for inference. Meanwhile, Dr.Fill, uses a modified depth-first search known as limited discrepancy search, as well as a post-hoc local search with heuristics to score alternate puzzles.\medskip

\noindent \textbf{Standalone QA Models for Crosswords} Past work also evaluated QA techniques using crossword question-answer pairs. These include linear models~\cite{barlacchi-etal-2014-learning}, WordNet suggestions~\cite{thomas2019crosswords}, and shallow neural networks~\cite{severyn-etal-2015-distributional,hill-etal-2016-learning}; we instead use state-of-the-art transformer models.\medskip

\noindent \textbf{Ambiguous QA} Solving crossword puzzles requires answering ambiguous and underspecified clues while maintaining accurate estimates of model uncertainty. Other QA tasks share similar challenges~\cite{watson,rodriguez2019quizbowl,rajpurkar2018squad2,min2020ambigqa}. Crossword puzzles pose a novel challenge as they contain unique types of reasoning and linguistic phenomena such as wordplay.\medskip

\noindent \textbf{Crossword Themes} We have largely ignored the presence of themes in crossword puzzles. Themes range from simple topical similarities between answers to puzzles that must be filled in a circular pattern to be correct. While Dr.Fill \cite{ginsberg2011dr} has a variety of theme handling modules built into it, integrating themes into our probabilistic formulation remains as future work.\medskip 

\noindent \textbf{Cryptic Crosswords} We solve American-style crosswords that differ from British-style ``cryptic'' crosswords~\cite{efrat2021cryptonite,rozner2021decrypting}. Cryptic crosswords involve a different set of conventions and challenges, e.g., more metalinguistic reasoning clues such as anagrams, and likely require different methods from those we propose.
\section{Conclusion}

We have presented new methods for crossword solving based on neural question answering, structured decoding, and local search. Our system outperforms even the best human solvers and can solve puzzles from a wide range of domains with perfect accuracy. Despite this progress, some challenges remain in crossword solving, especially on the QA side, and we hope to spur future research in this direction by releasing a large dataset of question-answer pairs.
In future work, we hope to design new ways of evaluating automated crossword solvers, including testing on puzzles that are designed to be difficult for computers and tasking models with puzzle generation.

\section*{Ethical Considerations}
Our data comes primarily from crosswords published in established American newspapers and journals, where a lack of diversity among puzzle constructors and editors may influence the types of clues that appear. For example, only 21\% of crosswords published in \textit{The New York Times} have at least one woman constructor~\cite{xwordwomen} and a crossword from January 2019 was criticized for including a racial slur as an answer~\cite{xwordracial}. We view the potential for real-world harm as limited since automated crossword solvers are unlikely to be deployed widely in the real world and have limited potential for dual use. However, we note that these considerations may be important to researchers using our data for question answering research more broadly.
\eric{we may want to mention copyright again, and what sort of license we are releasing the data under.}
\eric{may want to mention that no user data or PII is present in the clues.}

\eshaan{in section 2.1, we say our dataset is diverse in authorship yet provide evidence for the opposite here under ethical considerations?}
\section*{Acknowledgements}

We thank Sewon Min, Sameer Singh, Shi Feng, Nikhil Kandpal, Michael Littman, and the members of the Berkeley NLP Group for their valuable feedback. We are also grateful to Will Shortz and the organizers of the American Crossword Puzzle Tournament for allowing us to participate in the event. This work was funded in part by the DARPA XAI and LwLL programs. Nicholas Tomlin is supported by the National Science Foundation Graduate Research Fellowship.
\bibliographystyle{acl_natbib}
\bibliography{journal-abbrv,references}

\clearpage
\appendix
\section{Details of Qualitative Analysis}\label{appendix:qualitative}
In this section, we provide rough definitions for the categories used to construct Table~\ref{tab:clue_categories} and conduct the manual QA error analysis in Figure~\ref{fig:failure_categorization}:

\paragraph{Knowledge} Clues that require knowledge of history, scientific terminology, pop culture, or other trivia topics. Answers to knowledge questions are frequently multi-word expressions or proper nouns that may fall outside of our closed-book answer set, and clues often involve additional relational reasoning, e.g., \textit{Book after Song of Solomon} (\textsc{isaiah}).

\paragraph{Definition} Clues that are either rough definitions or synonyms of the answer.

\paragraph{Commonsense} Clues that rely on relational reasoning about well-known entities. These clues often involve subset-superset, part-whole, or cause-effect relations, e.g., \textit{Cause of a smudge} (\textsc{wetink}).

\paragraph{Wordplay} Clues that involve reasoning about heteronyms, puns, anagrams, or other metalinguistic patterns. Such clues are usually (but not always) indicated by a question mark.

\paragraph{Phrase} Clues or answers that involve common phrases or multi-word expressions. These clues are often written with quotation marks or blanks and their answers are frequently synonymous expressions, e.g., \textit{Hey man!} (\textsc{yodude}).

\paragraph{Cross-Reference} Clues that require knowledge of other elements in the puzzle, either through explicit reference (e.g., \textit{See 53-Down}) or due to their usage of crossword themes.

\section{Additional Dataset Statistics}\label{appendix:dataset}

\begin{figure}
\centering
\includegraphics[trim={0.5cm 0.5cm 0.5cm 0.5cm},clip,width=1.0\columnwidth]{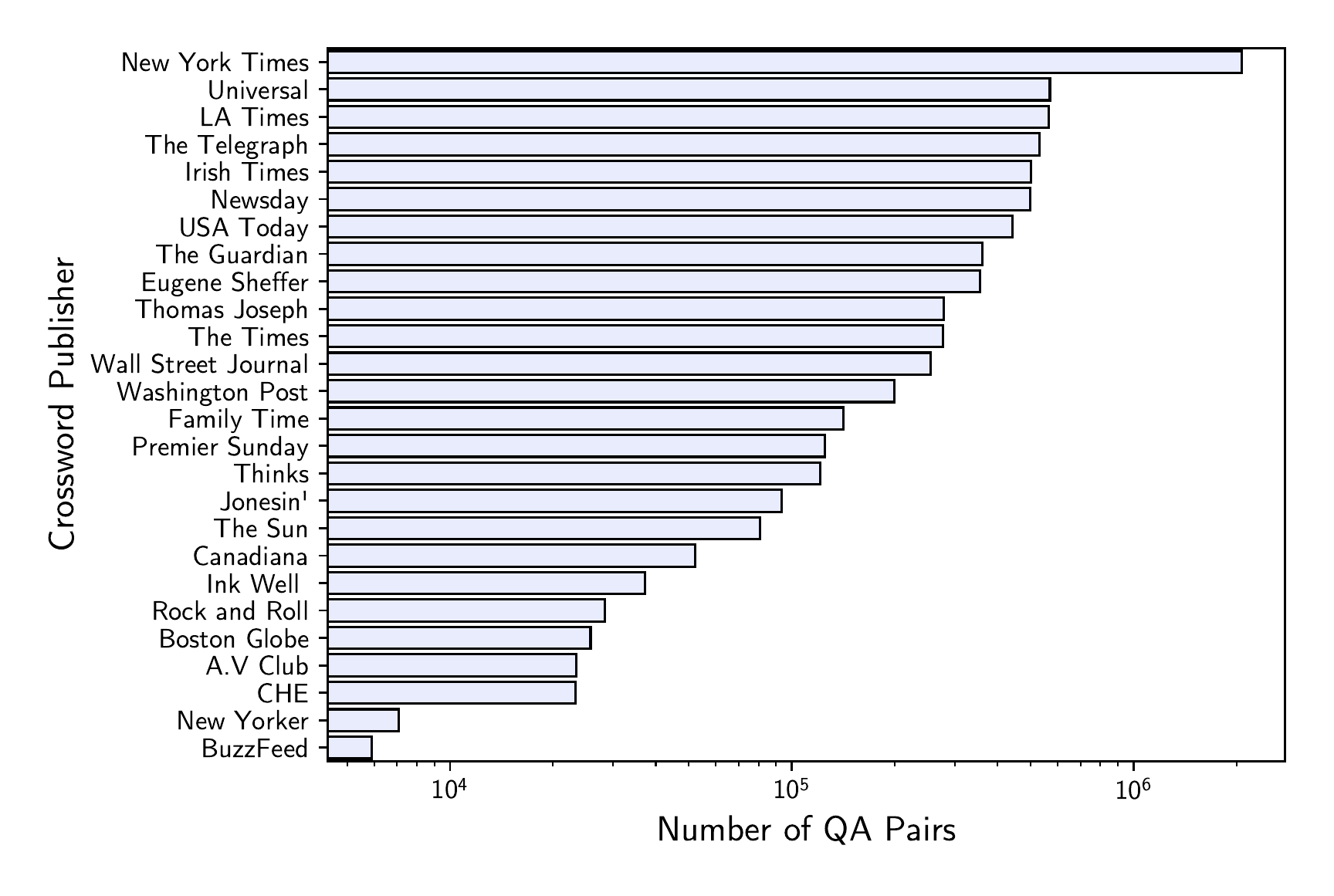}
\vspace{-0.73cm}
\caption{We build our dataset by collecting data from 26 publishers. Using a diverse set of publishers is beneficial as each publisher has different question types, answer distributions, and puzzle idiosyncrasies.}
\label{fig:publishers}
\end{figure}

\begin{figure}
\centering
\includegraphics[trim={0.5cm 0.5cm 0.5cm 0.5cm},clip,width=1.0\columnwidth]{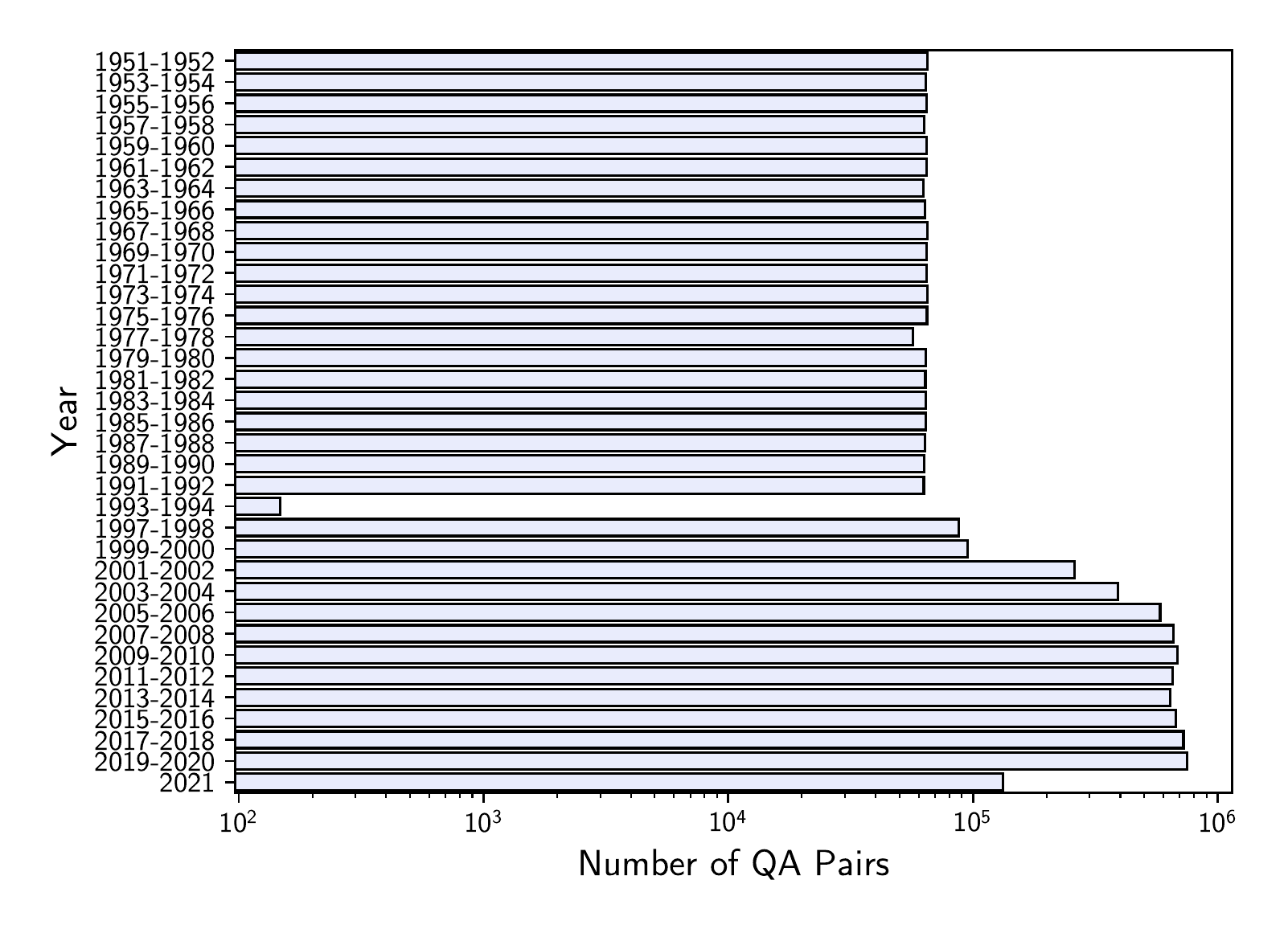}
\vspace{-0.73cm}
\caption{Our dataset spans over 70 years of crossword puzzles. The dip in puzzles in 1993-1996 is due to an unavailability of NYT puzzles from those years.}
\label{fig:years}
\end{figure}

\begin{figure}
\centering
\includegraphics[trim={0.5cm 0.5cm 0.5cm 0.5cm},clip,width=1.0\columnwidth]{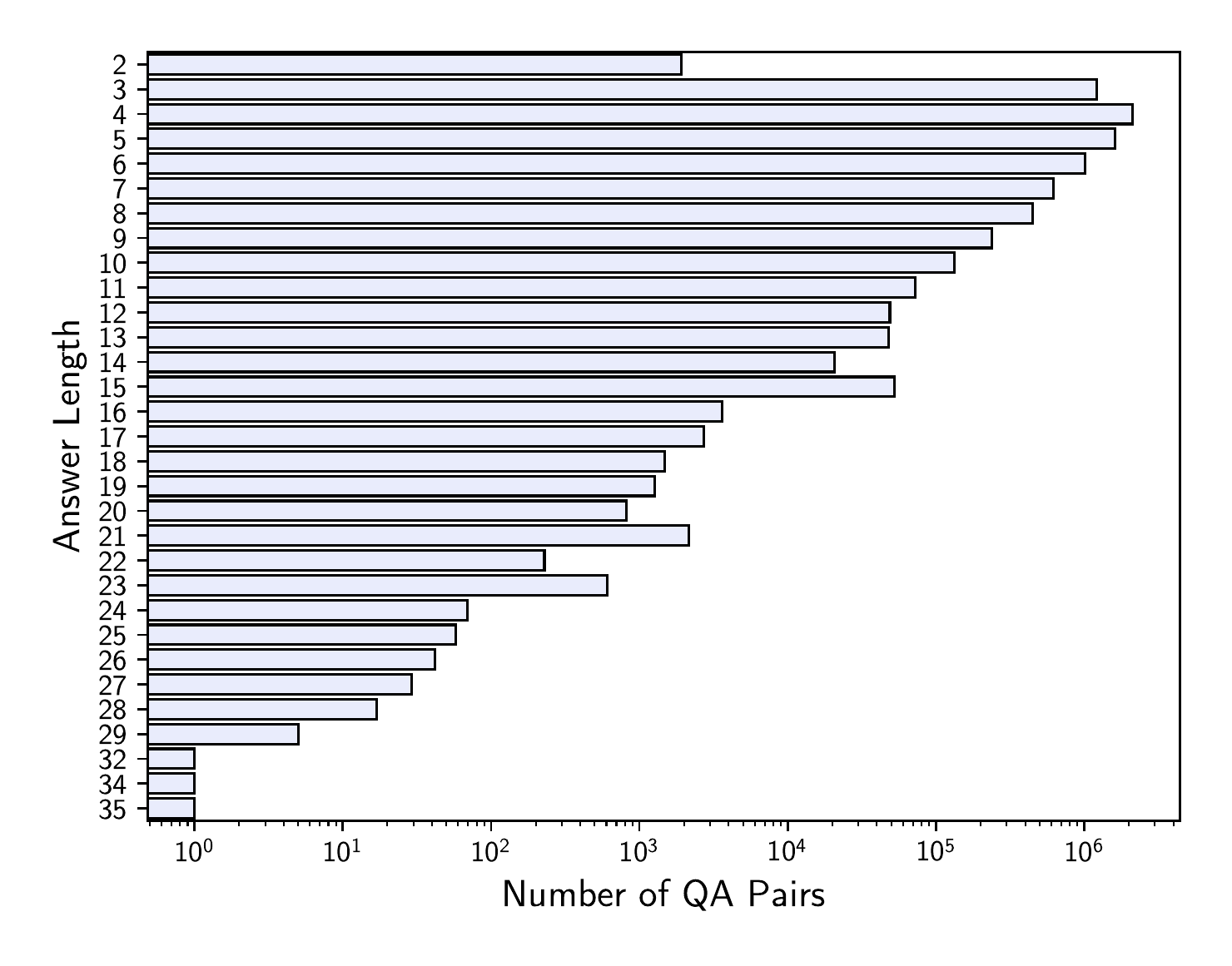}
\vspace{-0.73cm}
\caption{The answers in our dataset span many different lengths; longer answers are typically more difficult multi-word expressions or theme answers.}
\label{fig:answers}
\end{figure}

Figures~\ref{fig:publishers}--\ref{fig:answers} present a breakdown of the publishers, years, and answer lengths that are present in our crossword dataset.

\newpage
\section{\acpt{} Details}\label{appendix:acpt}

\begin{table}[h]
\centering
\begin{tabular}{lcrr}
\toprule 
\textbf{System} & \textbf{Year} & \textbf{Score} & \textbf{Rank} \\
\midrule 
Proverb & 1998 & 6,215 & 213th \\
Dr.Fill & 2012 & 10,060 & 141st \\ % TODO: check this 10060 number
Dr.Fill & 2013 & 10,550 & 92nd \\
Dr.Fill & 2014 & 10,790 & 67th \\
Dr.Fill & 2015 & 10,920 & 55th \\
Dr.Fill & 2016 & 11,205 & 41st \\
Dr.Fill & 2017 & 11,795 & 11th \\
Dr.Fill & 2018 & 10,740 & 78th \\
Dr.Fill & 2019 & 11,795 & 14th \\
\midrule
BCS QA + Dr.Fill & 2021 & 12,825 & 1st \\
BCS QA + BP + LS & 2021 & 13,065 & 1st \\
\bottomrule
\end{tabular}
\caption{Performance over the years in the \acpt{}. Dr.Fill has steadily improved due to system changes and increased training data. We also provide a retrospective evaluation of our final system (bottom row). Note that the 2020 ACPT was cancelled due to COVID-19.}
\label{tab:drfill_acpt}
\end{table}

\paragraph{Scoring System} The main portion of the \acpt{} consists of seven crossword puzzles. Competitors are scored based on their accuracy and speed. For each puzzle, the judges award:
\begin{itemize}[itemsep=0pt,leftmargin=10pt]
\item 10 points for each correct word in the grid,
\item 150 bonus points if the puzzle is solved perfectly,
\item 25 bonus points for each full minute of time remaining when the puzzle is completed. This bonus is reduced by 25 points for each incorrect letter but can never be negative.
\end{itemize}
The total score for the seven puzzles determines the final results, aside from a special playoff for the top three human competitors.
Table~\ref{tab:drfill_acpt} shows scores over the years for the \acpt{}, including our 2021 submission.

\section{Additional Analysis Results}\label{appendix:analysis}

\begin{figure}[t]
\centering
\includegraphics[trim={0.5cm 0.5cm 0.5cm 0.5cm},clip,width=\linewidth]{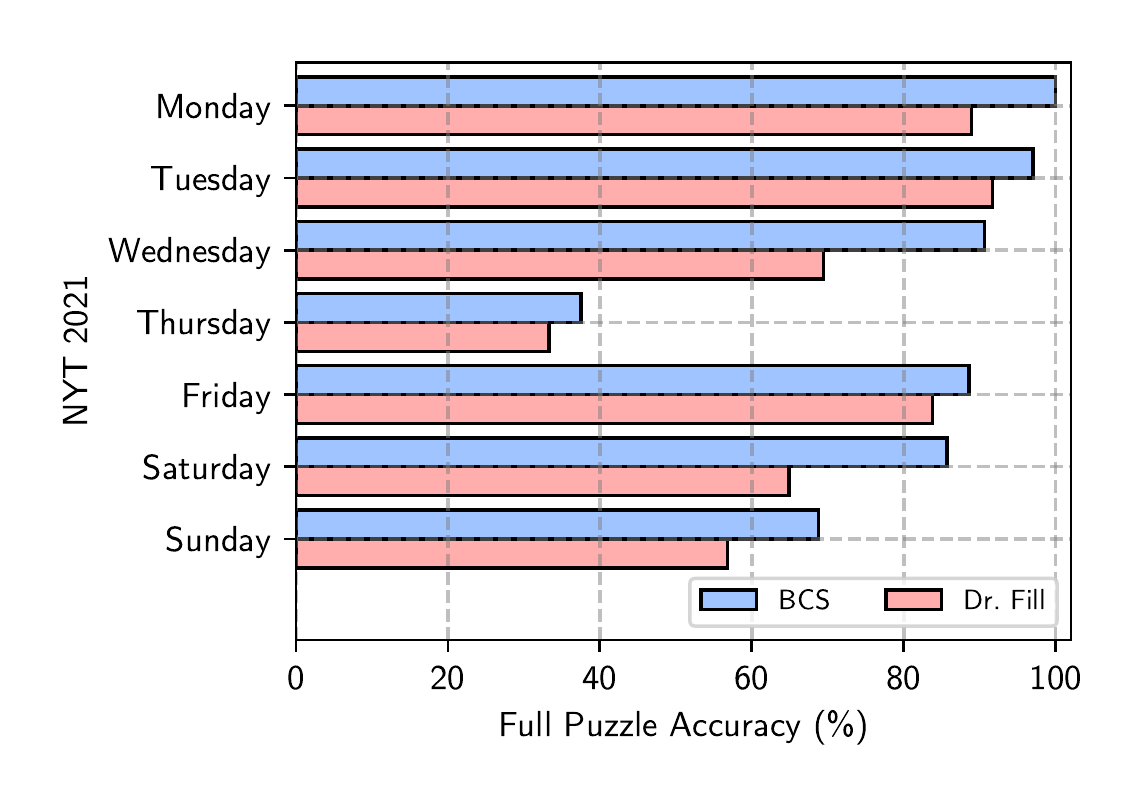}
\vspace{-0.70cm}
\caption{We compare our system's accuracy on NYT puzzles to the previous state-of-the-art Dr.Fill system and break down the results by day of the week. Both systems succeed on early week puzzles but struggle on Thursday puzzles that often contain unusual themes.}
\label{fig:puz_acc_vs_day}
\end{figure}

\begin{figure}[t]
\centering
\includegraphics[trim={0.8cm 0cm 0.6cm 0cm},width=\linewidth]{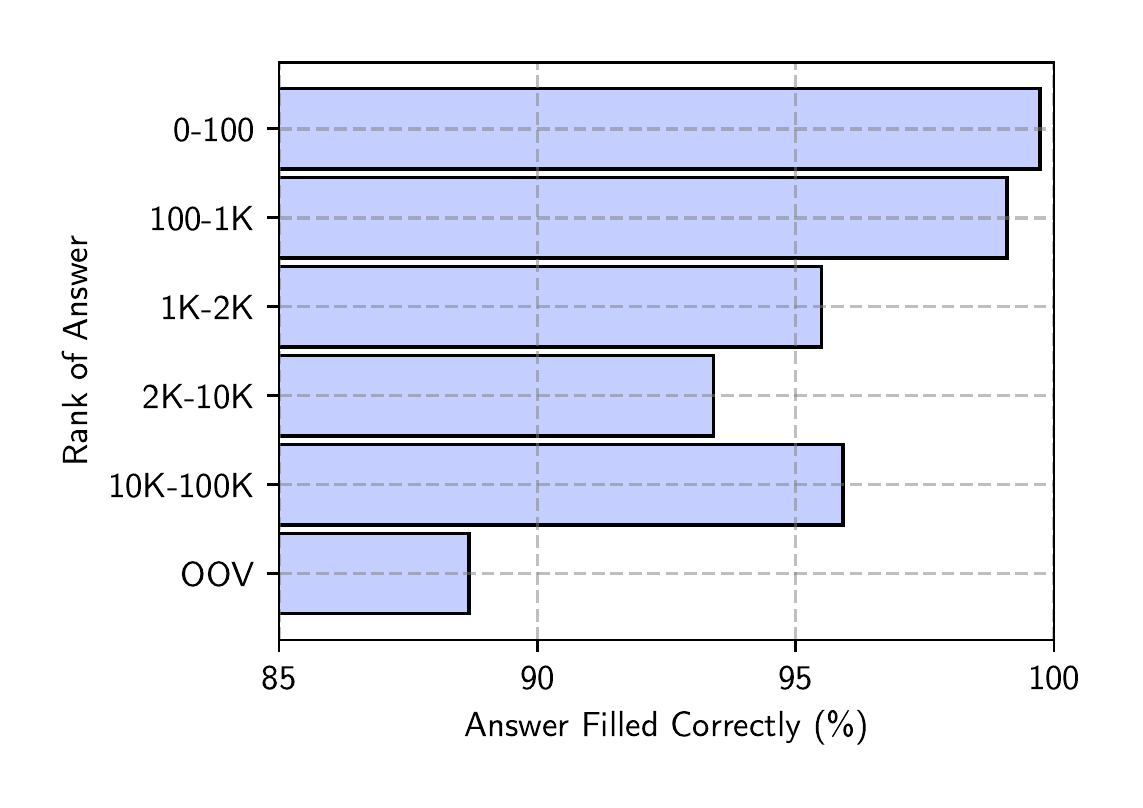}
\vspace{-1.1cm}
\caption{The chance that the BCS correctly fills in an answer as a function of the rank of the answer under its QA model. If the QA model predicts the answer in its top 1,000 candidates, it is usually filled in correctly.} 
\label{fig:unk_words_accuracy}
\end{figure}

Figure~\ref{fig:puz_acc_vs_day} shows our accuracy broken down by day of the week. Monday and Tuesday NYT puzzles---ones designed to be easier for humans---are also easy for computer systems. On the other hand, Thursday NYT puzzles, which often contain unusual theme entries such as placing multiple letters into a single grid, are the most difficult. Our system is unaware of these special themes, but the Dr.Fill system includes various methods to detect and resolve them and is thus more competitive on Thursday NYT puzzles. Finally, our system provides the largest gains on Saturday NYT puzzles which contain many of the hardest clues from a QA perspective.

We also compute results on themeless NYT puzzles. Themed puzzles range from topical similarity between answers in a puzzle, to multiple words ending with the same suffix, to multiple letters fitting inside a single square (i.e., rebus puzzles). For evaluation purposes, we consider themed puzzles to be any puzzle that contains a rebus\footnote{\url{https://www.xwordinfo.com/rebus}} or a circled letter\footnote{\url{https://www.xwordinfo.com/circles}} according to XWord Info, but this does not capture all possible themes.

% \begin{figure}[h]
% \centering
% \includegraphics[trim={0.3cm 0.3cm 0.3cm 0.3cm},clip,width=1.0\columnwidth]{figures/qa_acc_vs_fill_len.pdf}
% \vspace{-0.75cm}
% \caption{QA accuracy at top-1000 as a function of the fill length. Accuracy drops dramatically as the fill length increases, partially because longer answers contain more difficult multi-word expressions.}
% \label{fig:qa_acc_vs_fill_len}
% \end{figure}

\begin{figure*}[t]
%\captionsetup[subfigure]{}
\begin{subfigure}{0.46\textwidth}
\centering
\includegraphics[trim={0cm 0cm 0cm 0.3cm},clip,width=1.0\textwidth]{figures/ii0.png}
\caption{Before Local Search}
\end{subfigure}%
\hfill
\begin{subfigure}{0.46\textwidth}
\centering
\includegraphics[trim={0cm 0.35cm 0cm 0.05cm},clip,width=1.0\textwidth]{figures/ii1.png}
\caption{Step \#1}
\end{subfigure}%
\hfill
\begin{subfigure}{0.46\textwidth}
\centering
\includegraphics[trim={0cm 0.15cm 0cm 0.25cm},clip,width=1.0\textwidth]{figures/ii2.png}
\caption{Step \#2}
\end{subfigure}%
\hfill
\begin{subfigure}{0.46\textwidth}
\centering
\includegraphics[trim={0cm 0.35cm 0cm 0.05cm},clip,width=1.0\textwidth]{figures/ii3.png}
\caption{Step \#3}
\end{subfigure}%

\vspace{0.5cm}
\centering
\small
\begin{tabular}{llllll}
\toprule
\textbf{Clue} & \textbf{Gold} & \textbf{Before} & \textbf{Step 1} & \textbf{Step 2} & \textbf{Step 3}\\
\midrule
Beloved, in Arabic & \textsc{habib} & \textsc{hadid} & \textsc{hahib} &  \textsc{habib}  & \textsc{habib} \\[0.0ex]
Ill-advised opinions & \xword{badtakes} & \xword{hottakes} & \textsc{hodtakes} & \textsc{badtakes} & \textsc{badtakes}\\[0.0ex]
Feeling on a lo-o-ong car trip & \xword{borate} & \xword{derate} & \xword{berate} & \xword{berate} & \xword{borate} \\[0.0ex]
Italian herbal liqueur  & \xword{amaro} & \xword{amore} & \xword{amore}  & \xword{amare} & \xword{amaro} \\[0.0ex]
Not radical & \xword{modernists} & \xword{moternists} &  \xword{modernists} &  \xword{modernists} &  \xword{modernists}\\[0.0ex]
Long fur scarfs  & \xword{stoli} & \xword{stoga} & \xword{stola} & \xword{stola} & \xword{stoli} \\[0.0ex]
Outcome of a coin flip, e.g.,  & \xword{purechance} & \xword{puraagance} & \xword{puraahance} & \xword{purechance} & \xword{purechance}\\[0.0ex]
Choose randomly, in a way  & \xword{castles} & \xword{tastgas} & \xword{castlas} & \xword{castles} & \xword{castles}\\[0.0ex]
Like toreadors, again and again  & \xword{charade} & \xword{tharads} & \xword{charads}& \xword{charade}& \xword{charade}\\[0.0ex]
``Get 'em!''  & \xword{sic} & \xword{saa} & \xword{saa} & \xword{sac} & \xword{sic}\\[0.0ex]
Worrisome uncertainties  & \xword{bigifs} & \xword{bogies} & \xword{bigies} & \xword{bigins} & \xword{bigins} \\[0.0ex]
Like taxis and Julius Caesar, once  & \xword{hailed} & \xword{gaoled} & \xword{hailed} & \xword{hailed} & \xword{hailed} \\[0.0ex]
Immunologist Anthony  & \xword{fauci} & \xword{euaci} & \xword{eluci}& \xword{nauci} & \xword{nauci} \\[0.0ex]
Suffix with coward & \xword{ice} & \xword{ics} & \xword{ics} & \xword{ice} & \xword{ice} \\
\bottomrule
\end{tabular}
\vspace{-0.15cm}
\caption{\underline{Top:} We show a larger version of Figure~\ref{fig:ii_example}. \underline{Bottom:} The clues and associated answers after each step.}
\label{tab:ii_details}
\end{figure*}

\end{document}